\documentclass[sigconf]{acmart}
\usepackage{multirow}
\usepackage{graphicx}
\usepackage[linesnumbered, ruled]{algorithm2e}
\usepackage{algorithmic} 
\usepackage{amsmath}
\usepackage{amsthm}

\usepackage{amsfonts}
\usepackage{bm}

\definecolor{third}{RGB}{139,69,19}
\definecolor{second}{RGB}{176,23,31}
\usepackage{textcomp,booktabs}
\usepackage{xspace}
\usepackage{float}
\usepackage{subcaption} 
\usepackage{colortbl} 
\usepackage{xcolor}

\usepackage{soul}

\usepackage{comment}

\AtBeginDocument{%
  }

\copyrightyear{2025}
\acmYear{2025}
\setcopyright{acmlicensed}
\acmConference[CIKM '25] {Proceedings of the 34th ACM International Conference on Information and Knowledge Management}{ November 10--14, 2025}{Seoul, Republic of Korea.}
\acmBooktitle{Proceedings of the 34th ACM International Conference on Information and Knowledge Management (CIKM '25), November 10--14, 2025, Seoul, Republic of Korea}
\acmISBN{979-8-4007-2040-6/2025/11}
\acmDOI{10.1145/3746252.3761140}

\begin{document}

\title{MMiC: Mitigating Modality Incompleteness in Clustered Federated Learning}

\author{Lishan Yang}
\orcid{0009-0001-4795-9265}
\affiliation{%
  \institution{The University of Adelaide}
  \city{Adelaide}
  \country{Australia}}
\email{lishan.yang@adelaide.edu.au}

\author{Wei Emma Zhang}
\orcid{0000-0002-0406-5974}
\affiliation{%
  \institution{The University of Adelaide}
  \city{Adelaide}
  \country{Australia}}
\email{wei.e.zhang@adelaide.edu.au}

\author{Quan Z. Sheng}
\orcid{0000-0002-3326-4147}
\affiliation{%
  \institution{Macquarie University}
  \city{Sydney}
  \country{Australia}}
\email{michael.sheng@mq.edu.au}

\author{Lina Yao}
\orcid{0000-0002-4149-839X}
\affiliation{%
  \institution{CSIRO’s Data 61 and The University of New South Wales}
  \city{Sydney}
  \country{Australia}}
\email{lina.yao@data61.csiro.au}

\author{Weitong Chen}
\orcid{0000-0003-1001-7925}
\affiliation{%
  \institution{The University of Adelaide}
  \city{Adelaide}
  \country{Australia}}
\email{weitong.chen@adelaide.edu.au}

\author{Ali Shakeri}
\orcid{0009-0008-1739-8154}
\affiliation{%
  \institution{The University of Adelaide}
  \city{Adelaide}
  \country{Australia}}
\email{ali.shakeri@adelaide.edu.au}

\begin{abstract}
In the era of big data, data mining has become indispensable for uncovering hidden patterns and insights from vast and complex datasets. The integration of multimodal data sources further enhances its potential.
Multimodal Federated Learning (MFL) is a distributed approach that enhances the efficiency and quality of multimodal learning, ensuring collaborative work and privacy protection. However, missing modalities pose a significant challenge in MFL, often due to data quality issues or privacy policies
across the clients. 
In this work, we present MMiC, a framework for \underline{M}itigating \underline{M}odality \underline{i}ncompleteness in MFL
within the \underline{C}lusters. 
MMiC replaces partial parameters within client models inside clusters to mitigate the impact of missing modalities. Furthermore, it leverages the Banzhaf Power Index to optimize client selection under these conditions. Finally, MMiC employs an innovative approach to dynamically control global aggregation by utilizing Markowitz Portfolio Optimization. 
Extensive experiments demonstrate that MMiC consistently outperforms existing federated learning architectures in both global and personalized performance on multimodal datasets with missing modalities, confirming the effectiveness of our proposed solution. Our code released at https://github.com/gotobcn8/MMiC.

\end{abstract}





\maketitle
\section{Introduction}
Federated Learning (FL), introduced by McMahan et al. \cite{mcmahan2017communication}, enables multiple participants to collaboratively train a unified model without sharing raw data, thus preserving privacy. This paradigm is particularly well-suited for applications such as medical image analysis, connected vehicles, blockchain and so on\cite{kaissis2020secure, pokhrel2020federated, lu2019blockchain}. However, the absence of data sharing in FL introduces significant challenges, including model heterogeneity, communication overhead, and data distribution disparities. 

Multimodal Federated Learning (MFL) extends the FL paradigm to multimodal tasks, which leverage the complementary nature of multiple data modalities (e.g., text, images, and audio) to achieve superior performance on large-scale datasets. However, a common challenge in MFL is modal incompleteness, where certain modalities may be missing for some clients due to hardware limitations or data collection constraints. This issue can significantly degrade model performance, especially in heterogeneous scenarios. 
Existing approaches to tackling missing modalities primarily concentrate on two main aspects: the Data Processing Aspect, which involves generating absent modalities or their representations from the available ones, and the Strategy Design Aspect, which entails devising flexible model architectures capable of adapting to a variable number of available modalities during training, such as attention-based models~\cite{wu2024comprehensive}. 
While these methods have shown promise, they are often limited to specific models or datasets. For instance, Yu et al. \cite{yu2024robust} proposed a specialized LSTM-based model to handle incomplete modalities, but such solutions are not easily generalizable to other complex multimodal tasks because they rely heavily on the specific architecture
. Moreover, the lack of a unified approach to address modality incompleteness across diverse tasks and datasets further restricts their applicability in broader real-world scenarios.
In addition, the current MFL algorithm is limited to information exchange between clients and the server, lacking interaction between clients. For instance, PmcmFL \cite{baomissing2024} compensates for missing modality through server aggregation 
prototypes. However, due to constraints such as Non-IID data, the aggregated prototypes may not be suitable for all clients. 

Clustered Federated Learning (CFL), a specialized form of FL Personalization~\cite{tan2022towards}, groups clients into clusters based on models, parameters, or datasets. Collaborative training within each cluster reduces the impact of data distribution disparities between clients. Recent advancements in CFL, such as one-shot clustering methods like PACFL~\cite{vahidian2023efficient}, leverage the decomposability of data to identify client clusters efficiently. 
Each cluster inherently contains clients with similarities in both data and model parameters. We can leverage this characteristic of federated clustering to improve the training scenarios involving missing modalities. 
This means that \textit{the model parameters from clients within the same cluster that have full modalities can be used to fill in the parameters of clients with missing modalities}. 
Due to the constraints of communication efficiency within clusters, we can shrink replaced parameters to only one layer most significantly affected by the missing modalities. 

Current CFL algorithms predominantly employ random client selection~\cite{ghosh2019robust,ghosh2020efficient,fedsoft}, which ensures fairness over multiple rounds of updates by granting each client an equal opportunity to participate in aggregation. However, we do not expect clients with severe missing modalities to enjoy the same level of aggregation rights. Therefore, by analyzing performance variations across client training rounds, we can determine whether a client is frequently affected by missing modalities. This allows us to adjust the probability of client selection accordingly.

Since the global model, due to the intermediate aggregation process involving the cluster-client hierarchy, cannot directly perceive changes in clients' missing modalities, we can employ certain functions or metrics to evaluate the training process within each cluster. This evaluation yields a specific value that can be reported to the server, enabling the server to determine the aggregation weights for the current round.

To achieve these goals, we propose MMiC, a framework that integrates our solutions for the three issues. 
In particular, we employ the rate of parameter variation to identify the parameters to be replaced. Moreover, the parameters from clients possessing complete modality data and belonging to the same cluster will be utilized to supplant those of clients with incomplete modality data.
%
To enhance the current cluster client selection strategy in the presence of missing modalities, we incorporate the Banzhaf Power Index (BPI), a well-known coalitional game theory, to assess the contribution of each client. This approach enables us to preferentially select the clients who are more contributable, to perform more efficient and effective aggregation processes.
%
We further make use of Markowitz Portfolio Optimization, a return-risk balancing asset optimization algorithm to assess the training process on clusters in a nonlinear manner compared to the standard weighted averaging~\cite{2024avg}. The global model adjusts the weights of parameter updates based on the scores obtained from 
the clusters, thereby reducing the impact of missing modalities in local clients on the global model. 
MMiC is model-independent allowing for integration with various multimodal base models.
It is also modality-agnostic, enabling its application across diverse multimodal scenarios regardless of the specific modalities involved.
Our experiments demonstrate that MMiC significantly outperforms traditional FL and MFL algorithms in terms of both accuracy and stability. In summary, the main contributions of our work are as follows:
\begin{itemize}
    \item We propose a parameter substitution mechanism based on clustered federated learning, to mitigate the impact of missing modalities during training.
    \item We enhance the random client selection strategy by integrating the Banzhaf Power Index to favor clients with more significant.
    \item We are the first to apply the Markowitz Portfolio Optimization theory to evaluate the training process by incorporating the evaluation results into the global aggregation process to mitigate the negative impact of missing modalities on the global server.
\end{itemize}

\section{Related Works}
\label{sec:related works}
\subsection{Clustered Federated Learning (CFL)} 
Centralized federated learning trains a global model in a central server by aggregating gradients collected across numerous participating clients. 
Personalized federated learning (PFL) has sprung up designed to better suit the individual characteristics of each client to address the heterogeneity. PFL strategies like Multi-Task Learning, Personalization Layers and Clustered Federated Learning (CFL) have attracted considerable attention in the literature~\cite{smith2017federated,tan2022towards,mansour2020three,wu2020personalized}.

The fundamental concept of 
CFL is to group clients based on the similarity of their model parameters, data, or gradients. CFL emphasizes improving cooperation within these clusters to optimize the performance of cluster-specific models, while also enhancing overall global model performance.
Ghosh et al. \cite{ghosh2019robust} first introduced the architecture for CFL, and recommended clustering based on the similarity of model parameters. 
%
Fedsoft further advanced the iterative CFL by allowing a client to belong to multiple clusters, outperforming previous efforts. However, these methods incurred substantial resource consumption. 
One-shot clustering was then introduced by Dennis et al. \cite{kurian2021heterogeneity} for more efficient clustering. Their proposed method efficiently determined clustering attribution, identifying the uniqueness of client metrics within complex federated architectures.
%
Vahidian et al.~\cite{vahidian2023efficient} proposed a one-shot CFL algorithm based on data subspace clustering, They measured the similarity between clients by decomposing the dataset via Singular Value Decomposition (SVD). 
Our proposed MMiC framework is under the same setting as the one-shot CFL~\cite{vahidian2023efficient,yanglsh} and extends it to multimodal scenarios with missing modalities.
\subsection{Multimodal Federated Learning (MFL)}
Multimodal Federated Learning (MFL) 
extends standard federated learning to address multimodal datasets on clients and global server. 

\noindent \textbf{ Challenges and Advances in MFL}. 
FedCMR is the first to address multimodal tasks, though it encountered challenges such as modality discrepancies and missing data. Recent works have made strides in tackling these issues~\cite{zong2021fedcmr}. 
Graph-based FL represents a special case of CFL, they could be used to handle the heterogeneous modality fusion. FedMSplit~\cite{fedmsplit2022} employs a dynamic and multi-view graph structure to adaptively capture the correlations amongst multimodal local models. 
AGHG~\cite{AHGH2024} creates a global knowledge enhancer to extract information from different modalities.
These works collectively provide a foundation for understanding the challenges and solutions in MFL~\cite{mflsurvey}. 


\noindent \textbf{MFL with Missing Modality}.
CACMRN~\cite{10.1145/3581783.3611757} is the first FL framework to propose solutions for missing modalities. 
They aim to achieve modality \textbf{reconstruction} using existing multi-modal data by training multiple functionally diverse models. 
Yu et al.~\cite{yu2023multimodal} aim to address the issue of missing modalities in cross-modal retrieval. Clients, each with distinct modalities, learn the representation information of specific modalities and upload it to the server for \textbf{contrastive learning}. 
Fed-Multimodal~\cite{feng2023fedmultimodal} is a multimodal benchmark built upon the FL algorithm. The work presented a range of models and datasets and proposed a method to initialize the missing modal vector as \textbf{zero}. 
PmcmFL~\cite{baomissing2024} applied the common \textbf{prototype exchange} method of federated learning against Non-IID to the missing modality, and they store prototypes of all modalities for all classes from each client on the server. When encountering missing modalities, it utilizes the corresponding class-specific prototypes to fill in the representations.
In contrast, our approach does not impose restrictions on the types of multimodal tasks, meaning that we do not require the use of specific models. 

\section{Clustered Federated Learning}
\label{sec:cfl_base}
\textbf{Optimization Objective.} Given a global server $\mathcal{G}$ and $K$ clients $\{c_i\}^K_1$,  let $\mathcal{D}_k$ over $\mathcal{X} * \mathcal{Y}$  be the local dataset of  the $k$-th client $c_k$, with $\mathcal{X}$ and $\mathcal{Y}$ represent the data inputs and labels, $n_k = \vert \mathcal{D}_k \vert$ be the dataset size. Client aggregated weight is calculated as $w_k = \frac{n_k}{\sum_{k}^{K}n_k}$. $c_k$ and $\mathcal{G}$ possess the model $\theta_k$ and $\theta_{\mathcal{G}}$ respectively and the models are the same when initialized. In each round of federated training, on the client, it is to learn local model $\theta_k$ by minimizing the local empirical loss: $F_i(\theta_k) = \sum_{\xi \in \mathcal{D}_k } \frac{\ell(\theta_k,\xi)}{n_k} $. 
Where $\ell(\cdot,\cdot)$ is the pre-defined task-driven loss function and $\xi$ represents sampled training data in a batch. The clusters are defined by partitioning $K$ clients into $M$ disjoint subsets. 
Different from traditional federated selection, clustered federated selection is refined to each cluster, and each round has client subset $\mathcal{S}_m$ selected in $\mathcal{M}_m$. 
The weight of cluster $m$ is usually calculated as $w_m = \frac{\sum_{k \in \mathcal{M}_m}n_k}{\sum_{1}^{M}\sum_{k \in \mathcal{M}_m}n_k}$. Therefore, the global objective function can be expressed as:
\begin{equation}
\begin{alignedat}{2}
 \quad min \quad \sum_{m=1}^{M} \sum_{k \in \mathcal{M}_m} w_k \ell(\boldsymbol{\theta}_{k})
\end{alignedat}
\label{eq:global}
\end{equation}

\vspace{1mm}
\noindent \textbf{Model Aggregation.} 
We obtain cluster models from clients' models by using \textbf{FedAvg}~\cite{li2019convergence}: $ \boldsymbol{\Theta}_m = \sum_{k \in \mathcal{M}_m} w_k \boldsymbol{\theta}_{k}$. Our method for aggregating cluster models into the global model is elaborated in Section~\ref{sec:method_porfolio}.

\section{Methodology}
\label{sec:method}
\begin{figure}[tb]
\centering
  \includegraphics[width=\linewidth]{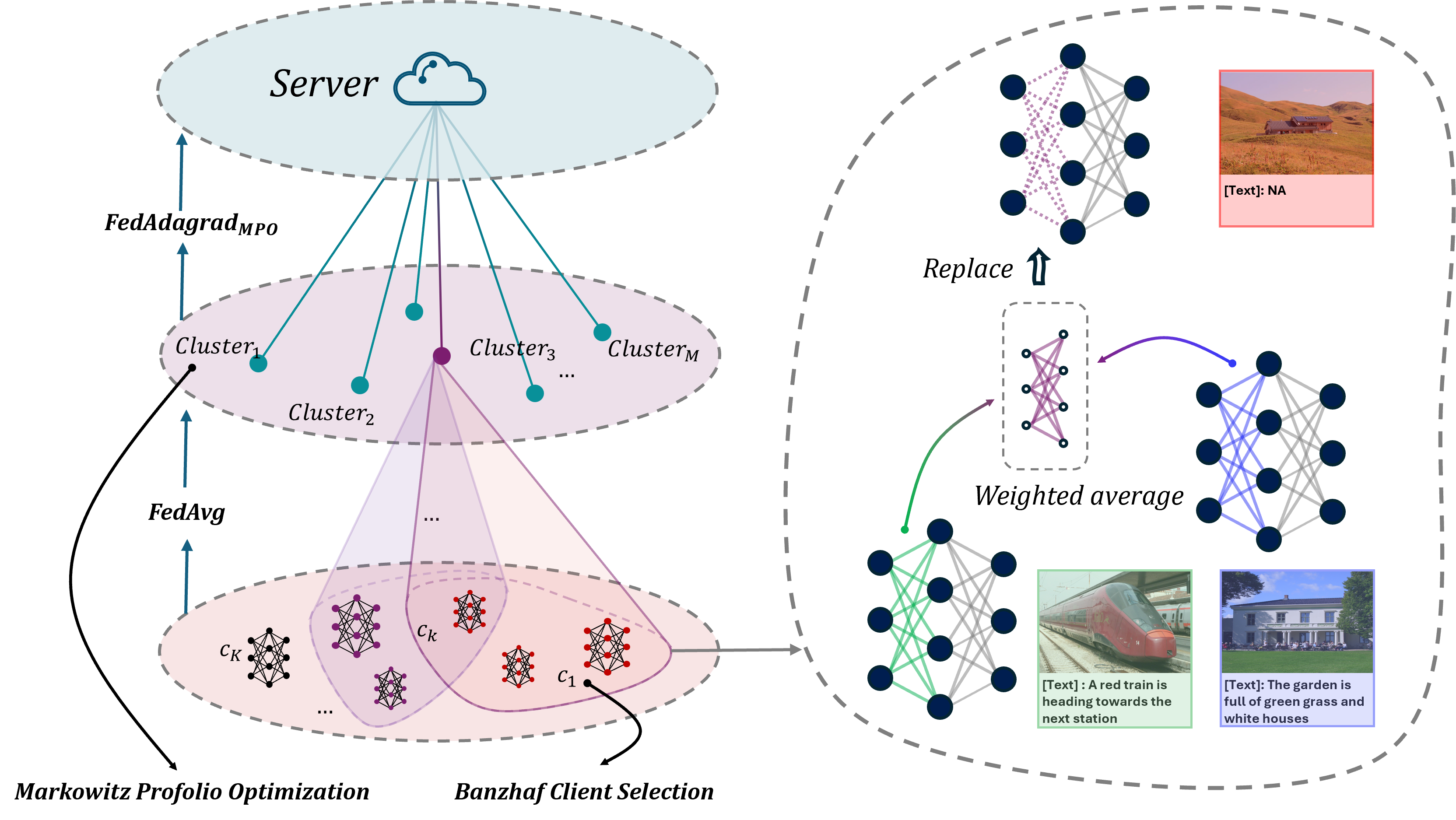}
  \caption{The MMiC architecture is divided into three layers: Server, Cluster, and Client. 
  The right part illustrates the process of parameter substitution within a cluster.
  The MPO module primarily operates during the aggregation process from clusters to the global model, while the Banzhaf client selection is applied to the client selection within each cluster.
  }
  \Description{}
  \label{fig:architecture}
\end{figure}

Figure~\ref{fig:architecture} 
depicts the architecture of MMiC. 
The left part illustrates the overall architecture including the clients at the bottom, clusters in the middle, and the global server at the top. 

Each cluster accommodates multiple clients, and each client hosts multimodal data with missing modalities locally. The cluster model is derived from the federated average aggregation (Section~\ref{sec:cfl_base}). 
Within each cluster, MMiC handles the missing modality issue and the client selection task. First, a parameter layer substitution mechanism allows clients with missing modalities to obtain knowledge from clients with complete modality data within the same cluster (Section~\ref{sec:method_mm}). Second, client selection is performed leveraging the Banzhaf Power Index, which identifies a set of \textit{core} clients that contribute the most effectively to the cluster's model (Section~\ref{sec:method_banzhaf}). 

For global aggregation, MMiC extends $FedAdagrad$~\cite{reddi2020adaptive} 
by adopting the Markowitz Portfolio Optimization 
(Section~\ref{sec:method_porfolio}). 
This method incorporates client training evaluation results from each cluster and reduces the impact of missing modalities by down-weighting the contribution from clusters that contain missing modalities.


\subsection{Cluster-based Parameter Substituion}
\label{sec:method_mm}


We consider a multimodal setting where $i$-th data instance is a pair $\{X^{text}_i, X^{image}_i\}$. Missing modalities occur randomly, yielding samples such as $\{X^{null}_i, X^{image}_i\}$ or $\{X^{text}_i, X^{null}_i\}$\footnote{Please note that MMiC is modality-agnostic -- it remains effective regardless of the type or number of modalities, as long as modalities are consistent across clients. We have additional experiments on other modalities shown in Section \ref{apx:othermodality}}.

To mitigate the consequences of missing modalities in each client, parameters that are less impacted by modality incompleteness are used to substitute those that are more significantly affected. We first propose a substitution strategy that can assist these clients in the training rounds. This strategy is inspired by the CFL setting, where a cluster model represents an aggregation of client models within the same cluster. The core idea is to substitute the parameters of client models with missing modality data with those of clients in the same cluster that have complete modality data. This approach involves two key tasks: (i) identifying the parameters to be substituted and (ii) ensuring the efficiency of the method.

\vspace{1mm}
\noindent \textbf{Poverty Parameter Identification.} 
In one round of client model training, we identify the layer that is the most affected by missing modalities as $poverty$ layer. 
We use the parameter difference between the last batch $b$-1 and the current batch $b$ to define the degree of this effect when the missing modality occurs in batch $b$ and batch $b-1$ is with complete modality\footnote{We are aware of the missing modality status of the data.}:
\begin{equation}
    \rho_{k,t}^{e,b} (p) = mean\left(
    \left|\frac{ \boldsymbol{\theta}_{k,t}^{e,b} (p)- \boldsymbol{\theta}_{k,t}^{e,b-1} (p) }{\boldsymbol{\theta}_{k,t}^{e,b-1}(p)} \right|
    \right) \; 
 \text{and} \; \rho_{k,t}^{e,b} (p) \in \mathbb{R}
\end{equation}
%
%
where $e$ represents the epoch, $t$ is the training round, $k$ denote the client. $\rho_{k,t}^{e,b} (p)$ represents average parameter change rate of the $p$-th layer.
As $\rho_{k,t}^{e,b}$ is a set containing all layers' change rates. 
We collect all the batches in the $e$-th epoch that contain complete modality data, while the preceding batch contains missing modality data, and form a batch set $\mathcal{B}*$. Therefore, the relative change rate for the layer $p$ in one epoch could be expressed as:
\begin{equation}
\label{eq:param_change_p}
\rho_{k,t}^{e} (p) = \frac{ \sum_{b \in \mathcal{B}^*} \rho_{k,t}^{e,b} (p)}{|\mathcal{B}^*|}
\end{equation}

After calculating the average change rate for all layers in the model in one epoch, we identify the layer with the highest change rate across all epochs. This layer would be considered as the $poverty$ layer with index $\mathcal{P}_{k,t}$:
\begin{equation}
    \mathcal{P}_{k,t} =  \arg\max_{\substack{p \in \{1, 2, \dots, P\}}} \rho_{k,t} (p)
\end{equation}

It is worth mentioning that we select parameters from only one layer to reduce communication consumption.

\vspace{1mm}
\noindent \textbf{Poverty Parameter Substitution (PPS).} 
After identifying the poverty parameter in $c_k$, \textit{we propose to replace these parameters with the parameters from other clients in the cluster that have complete modality data}. The proposed parameter substitution can be considered as a model editing strategy~\cite{li2024pmet}. 
Specifically, we compute a weighted average of the corresponding parameters of other clients and replace the poverty parameter in $c_k$.
\begin{equation}
    \label{eq:param_sub}
    \boldsymbol{\theta}_{k,t}(\mathcal{P}_{k,t}) = \sum_{j = 1}^{J} {w_j}\boldsymbol{\theta}_{j,t}(\mathcal{P}_{k,t})\quad where\quad \mathcal{P}_{j,t} = \varnothing, K + J = |\mathcal{M}_m|
\end{equation}
where $c_k$ represents the poverty client and $c_j$ is the client that trained with the complete modality round, and both $c_k$ and $c_j$ belong to the cluster $\mathcal{M}_m $. 

\vspace{1mm}
\noindent \textbf{Optimizers Could Approximate Parameter Difference.} 
Saving parameters between two batches is resource-intensive and time-consuming. However, we can roughly deduce the rate of change by utilizing the gradient derivation function of the optimizer. Below, we provide the calculation process for parameter variations during the Adam optimization processes. Here $\hat{\boldsymbol{momentum}}$ is the bias-corrected moment estimate and $\hat{\boldsymbol{\nu}}$ is the second-moment estimate in Adam optimizer~\cite{kingma2014adam}:
\begin{align}
\label{eq:adam_efficiency}
    \Delta\boldsymbol{\theta}_{k,t}^{e,b} =  \frac{\eta_{k,t} abs(\-\hat{\boldsymbol{momentum}}_b)}{abs(\boldsymbol{\theta}_{k,t}^{e,b}) (\sqrt{\hat{\boldsymbol{\nu}}_b} + \epsilon)}  
\end{align}


\subsection{Banzhaf Client Selection in Clusters}
\label{sec:method_banzhaf}

The standard procedure for current CFL involves randomly selecting a subset of clients $\mathcal{S}_m$ from each cluster for training in each round, followed by a global aggregation. While the random selection is considered fair, there is a possibility that the cluster models aggregated by the selected clients may yield subpar results if some of the selected clients have a significant number of data instances with missing modalities. The negative impact will be transited to the global model through aggregation. 
Moreover, the negative impact will be enlarged over successive rounds of client training and cluster model aggregation, further degrading the performance of the global model.

Few CFL work addresses this issue and quantifies the impact of client selection for each cluster.
We propose a novel method to fill the gap by leveraging Banzhaf Power Index~\cite{dubey1979mathematical, laruelle2001shapley} (BPI) to select the core clients. 
It comprises two main steps. 
First, we identify core members within each subset selected in each round and tally the times of core members, denoted as \textit{core times}. 
Then, by leveraging the historical selection times and core times, we devise a new client selection probability formula, which is applied 
in our subsequent client selection process. Banzhaf client selection does not require extra communication. 

\vspace{1mm}
\noindent\textbf{Core Members Identification.}
%
At training round $t$, each cluster $\mathcal{M}_m$ makes a random client selection. We can obtain the $i$-th client's local model performance  $a_{i,t}$ (i.e., accuracy, F1 score or RSum) of each client in the $t$-th round, and the performance $a_{i,t-1}$  in the ($t$-1)-th round. We calculate the return $\alpha_{i,t}$ of the client for the $t$-th round based on the difference between these two rounds of performances as follows: 
\begin{equation}
\label{eq:alpha}
    \alpha_{i,t} = a_{i,t} - a_{i,t-1}
\end{equation}
where $\alpha_{i,t} < 0$ indicates that the client is generating a negative return to the \textit{investment portfolio}; conversely, $\alpha_{i,t} >0$ signifies a positive return. Then, the cluster's return is calculated as the sum of the client returns:
\begin{equation}
\label{eq:sum_alpha}
    A_{\mathcal{S}_{m,t}} = \sum_{i=0}^{|\mathcal{S}_{m,t}|} w_i \alpha_{i,t}
\end{equation}

For each cluster $\mathcal{S}_{m,t}$, we have its total return $A_{\mathcal{S}_{m,t}}$ by Equation~(\ref{eq:sum_alpha}). Correspondingly, we can calculate the total return for $\mathcal{S}_{m,t}^{'}$ when excluding client $c_i$, 
$\mathcal{S}_{m,t}' = \mathcal{S}_{m,t} \textbackslash \{i\}$: 
Then, we use the 
return of cluster model 
to evaluate the core members:
\begin{equation}
    c_i\, is\, core\, member?
    \begin{cases} 
        True &  A_{\mathcal{S}_{m,t}} \geq \alpha^m_t  \text{ and } A_{\mathcal{S}_{m,t}^{'}} < \alpha^m_t \\
        False & \text{otherwise}
    \end{cases}
\end{equation}
The above equation describes the process of determining whether client $c_i$ is a core member during the $t$-th round of client random selection. We 
utilize a \textit{counter} \textbf{$\phi(*)$} that starts with 0 to tally the number of times each client has been identified as a core member:
\begin{equation}
\label{eq:core_times}
        \phi(i) = \sum_{t=1}^{T} \mathbb{I}\left(c_i\, is\, a\, core\, member\, \right)
\end{equation}
where $\mathbb{I}(*)$ is 
the indicator function that returns 1 if the condition $*$ is satisfied, otherwise returns 0. 

\vspace{1mm}
\noindent \textbf{Client Selection Probability.} 
BPI prioritizes clients without missing modalities, thus favoring those that contribute more to performance improvements. We could design a method to measure the probability of a client being selected in the current round based on its core times. 
Our goal is to ensure that a higher core time corresponds to a higher probability. However, 
this way will make clients with small core times unlikely to be selected later. 
To solve this problem, we set another counter $\mathcal{T}(*)$ that records the number of times the client has been selected in its history, and we determine the final selection probability by balancing the historically selected times $\mathcal{T}(i)$ and the core times $\phi(i)$: 
\begin{align}
\label{eq:new_prob}
    prob(i) &=  \frac{e^{\frac{\tau \phi(i)}{\mathcal{T}(i)}}}{\sum_{i \in \mathcal{M}_m} e^{\frac{\tau \phi(i)}{\mathcal{T}(i)}}} 
\end{align}
For client $i$ in cluster $\mathcal{M}_m$, its probability in the $t$-th round to be selected is 
$prob(i)$. We increase the disparity in probability allocation for each client through an exponential function, where $\tau$ is a temperature constant. As $\tau$ increases, the disparity in probability increases, and as $\tau$ decreases, the disparity in probability decreases.

\subsection{Markowitz Portfolio Optimization for Global Aggregation}
\label{sec:method_porfolio}
%

In our framework, MMiC builds upon $FedAdagrad$~\cite{reddi2020adaptive} to support adaptive cluster-level aggregation in the presence of modality incompleteness. To enhance global robustness, we propose a novel integration of Markowitz Portfolio Optimization (MPO) into $FedAdagrad$. 

Originally introduced in financial theory~\cite{durand1960portfolio}, MPO constructs investment portfolios that trade off expected \textbf{return} and associated \textbf{risk}. Inspired by this principle, we adapt the concept to dynamically regulate the influence of each cluster in global aggregation. Specifically, clusters consisting of underperforming clients—such as those suffering from missing modalities—may introduce biased or unreliable updates. We treat each cluster as an ``investment'' and aim to down-weight risky clusters while preserving beneficial contributions.

Our approach introduces a scoring function that quantifies the overall quality of a cluster's update in each round, based on both its expected return and internal risk. This score directly informs the global aggregation process, enabling us to balance between current update quality and accumulated historical knowledge. 
To achieve this, we require three core indicators to construct our ``\textit{investment portfolio}'' for cluster $\mathcal{M}m$ includes the expected return (i.e., payoff) of the cluster, denoted as $A_{m,t}$ (Equation~\ref{eq:sum_alpha}); the risk $\bar{\sigma}_{m,t}$, derived from the intra-cluster covariance; and the return-risk tolerance coefficient $\lambda$ that governs the trade-off between return and risk for the cluster.

\vspace{1.5mm}
\noindent\textbf{Cluster Risk.} 
According to the asset risk calculation formula from MPO~\cite{durand1960portfolio,markowitz1991foundations}, we obtain the covariance as follows:
\begin{equation}
\label{eq:cov_risk}
    \bar{\sigma}_{m,t} = \sum_{i = 1}^{|\mathcal{S}_{m}|} \sum_{j = 1}^{|\mathcal{S}_{m}|} w_i w_j \left\{\frac{1}{T-1} \sum_{t = 1}^{T}(\alpha_{i,t} - \bar{\alpha}_{i,t-1})(\alpha_{j,t} - \bar{\alpha}_{j,t-1})\right\}
\end{equation}
where $\bar{\alpha}_{i,t-1}$ represents the historical average return for client $i$ in round $t-1$. $w_i$ and $w_j$ are weights assigned to the $i$-th and $j$-th clients respectively (Please refer to Section~\ref{sec:cfl_base} for weights' calculation). Then the Risk-Adjusted Return Coefficient (\textbf{RARC}) is formulated as follows:
\begin{equation}
\begin{aligned}
 \label{eq:mpo_eq}
     \quad \gamma_{m,t} = \lambda \bar{\sigma}_{m,t} - (1 - \lambda) A_{m,t}
\end{aligned}
\end{equation}
where $\lambda$ [0,1] is the risk tolerance constant. We defaultly set it as 0.5 to balance risk and return.
Section~\ref{apx:time_analysis} demonstrates how we reduce the cubic computational complexity of MPO to a quadratic complexity.

\vspace{3mm}
\noindent\textbf{Participate in Global Aggregation.} 
We have evaluated the training RARC $\gamma_{m,t}$ for each cluster. Next, we dynamically incorporate these RARC into the global aggregation process. A higher $\gamma_{m,t}$ indicates that the cluster has experienced a relatively poor local training process.
In such scenarios, our objective is to diminish the impact of the current round's parameter updates on the global aggregation by down-weighting the cluster parameters associated with higher $\gamma_{m,t}$.

To accomplish dynamic aggregation between current round and historical rounds, $FedAdagrad$ employs the following update mechanism: $\boldsymbol{\Delta_{m,t}}$ is the difference between the current cluster model and the global model. $\boldsymbol{mtn}_{m,t}$ is updated iteratively and collects the historical parameter updates, while $\beta$ balances the historical parameter updates $\boldsymbol{mtn}_{m,t-1}$ and $\boldsymbol{\Delta}_{m,t}$. A larger $\beta$ indicates the retention of more historical parameter information. $\boldsymbol{\upsilon}_t$ is an accumulated squared change estimate used to measure changes in parameters and adjust the learning rate to improve the stability and convergence speed of the optimization process: 
\begin{equation}
\begin{aligned}
& \boldsymbol{\Delta_{m,t}} =  w_m(\boldsymbol{\Theta}_{m,t} - \boldsymbol{\Theta}_{\mathcal{G},t-1}) \\
& \boldsymbol{mtn}_{m,t} = \beta \boldsymbol{mtn}_{m,t-1} + (1 - \beta) \boldsymbol{\Delta_{m,t}}, \beta \in [0,1)  \\
& \boldsymbol{\upsilon}_{m,t} = \boldsymbol{\upsilon}_{m,t-1} + \boldsymbol{\Delta}_{m,t}^2 \\
& \boldsymbol{\theta}_{\mathcal{G},t} = \boldsymbol{\theta}_{\mathcal{G},t-1} + \eta \frac{\boldsymbol{mtn}_{m,t}}{\sqrt{\boldsymbol{\upsilon}_{m,t}} + \tau}
\label{fm:fedadagrad}
\end{aligned}
\end{equation}
We introduce a dynamic $\beta_{m,t}^*$ to replace the static $\beta$ in $FedAdagrad$, which effectively captures the influence of $\gamma_{m,t}$: 
\begin{equation}
\begin{aligned}
    &\beta_{m,t}^* = \beta + \frac{tanh(ReLU(\gamma_{m,t}))}{\tau}\\
    & \boldsymbol{mtn}_{m,t} = \beta_{m,t}^* \boldsymbol{mtn}_{m,t-1} + (1 - \beta_{m,t}^*) \boldsymbol{\Delta_{m,t}}, \beta_{m,t}^* \in [0,1)  \\
\end{aligned}
\end{equation}
When $\gamma_{m,t}>0$, then $\beta_{m,t}^*>\beta$, which helps retain more historical parameter updates in the global aggregation. Otherwise, $\beta_{m,t}^*=\beta$.  
Here \textit{$\tau$} serves as the smoothness control constant for the range. 
We suggest setting the temperature to range [10,20], and it depends on the value set for $\beta$.

\subsection{Variants of MMiC}
\label{sec:method_mmic}

We propose two MMiC variants by extending two existing methods that address modality incompleteness. Specifically, by integrating these methods with the three proposed modules in Section \ref{sec:method_mm} - \ref{sec:method_porfolio}, we form the two MMiC variants.

\vspace{1mm}
\noindent\textbf{MMiC (Modality Reconstruction).} 
In this variant, we adopt the reconstruction idea from ~\cite{Sanmi2022} and incorporate two modality reconstruction models, Text-to-Image and Image-to-Text, at each cluster to reconstruct the missing modality using the data from the complete modality.  

\vspace{1mm}
\noindent\textbf{MMiC (GMC-Fusion).} The other MMiC variant is based on the GMC-Fusion model~\cite{poklukar2022geometric}. 
This variant necessitates the additional projecting of information from two modalities into the shared space, followed by computing the  Geometric Multimodal Contrastive loss using all available modality representations and shared space representations from a batch. 
In this way, the missing modality information could be filled in the shared space. 

\section{Experiments}
\subsection{Experiment Setup}
\label{sec:exp}
\begin{table*}[t]
    \centering
    \renewcommand{\arraystretch}{1.2} 
    \caption{Global and personalized performance compared with different algorithms. Bold with * indicated the best performance, while \textbf{bold black} indicated excellent or stable top three performance. Since the PmcmFL algorithm is only for classification tasks, it can't be compared in Flick30k.
    MR stands for Modality Reconstruction and GF is GMC-Fusion.
    }
    \resizebox{\textwidth}{!}{

    \begin{tabular}{lccc | ccc | ccc}
        \toprule
        \textbf{Global} & \multicolumn{3}{c}{\textbf{CrisisMMD (F1 score)}} & \multicolumn{3}{c}{\textbf{UPMC-Food101 (Accuracy)}} & \multicolumn{3}{c}{\textbf{Flickr30k (RSum score)}} \\
        & \multirow{2}{*}{mm@0.2} & \multirow{2}{*}{mm@0.2} & \multirow{2}{*}{mm@0.5} & \multirow{2}{*}{mm@0.2} & \multirow{2}{*}{mm@0.2} & \multirow{2}{*}{mm@0.5} & \multirow{2}{*}{mm@0.2} & \multirow{2}{*}{mm@0.2} & \multirow{2}{*}{mm@0.5} \\
        & mc@0.3 & mc@0.5 & mc@0.5 & mc@0.3 & mc@0.5 & mc@0.5 & mc@0.3 & mc@0.5 & mc@0.5 \\
        
        \midrule
        FedAvg [2017] & 21.886 ± 2.87 & 20.531 ± 1.9 & 24.095 ± 0.178 & 88.053 ± 1.03 & 87.626 ± 1.34 & 88.283 ± 0.83 & \textbf{14.324} ± 1.68 & 4.704 ± 0.38 & 3.812 ± 0.16 \\
        FedOpt [2021] & \textbf{29.320} ± 1.5 & \textbf{29.750} ± 2.04 & 26.292 ± 1.91 & 88.270 ± 0.53 & \textbf{89.064} ± 1.09 & 88.324 ± 0.85 & \textbf{16.138} ± 1.06 & 5.044 ± 0.61 & \textbf{12.576} ± 1.28 \\
        FedSoft [2022] & 11.371 ± 0.02 & 8.180 ± 0.12 & 10.464 ± 0.01 & 87.471 ± 0.38 & 85.984 ± 0.79 & 81.461 ± 0.14 & 1.189 ± 0.01 & 1.297 ± 0.01 & 1.106 ± 0.02 \\
        PACFL [2023] & 23.239 ± 3.35 & 20.274 ± 2.18 & 23.131 ± 3.029 & 88.152 ± 1.69 & 88.343 ± 1.31 & 88.439 ± 0.91 & 10.168 ± 1.14 & 8.013 ± 0.79 & 3.678 ± 0.25 \\
        CreamFL [2023] & 22.766 ± 0.09 & 18.131 ± 1.0 & 16.958 ± 2.93 & 87.539 ± 0.26 & 87.102 ± 0.17  & 87.390 ± 0.88 & 13.965 ± 0.04 & \textbf{13.857} ± 0.06 & 11.839 ± 0.02 \\
        PmcmFL [2024] & 24.610 ± 1.49 & 22.471 ± 0.44  & 22.946 ± 0.89 & 87.916 ± 0.90 & 85.159 ± 0.13 & 86.873 ± 0.52 & - & - & - \\
        \midrule
        \textbf{MMiC} &*\textbf{{29.462}} ± 0.36 & 27.815 ± 1.44 & *\textbf{{28.480}} ± 1.86 & \textbf{88.777} ± 0.86 & *\textbf{{89.436}} ± 1.21 & *\textbf{{89.428}} ± 0.78 & \textbf{14.266} ± 1.20 & 8.322 ± 0.177 & \textbf{14.220} ± 1.28 \\
        \textbf{MMiC (MR)} & \textbf{29.108} ± 0.86 & *\textbf{{30.471}} ± 0.77 & \textbf{28.354} ± 1.36 & *\textbf{{88.997}} ± 0.33 & \textbf{88.519} ± 2.31 & \textbf{89.044} ± 0.76 & 10.907 ± 1.21 & \textbf{10.152} ± 0.87 & 8.614 ± 0.84 \\
        \textbf{MMiC (GF)} & 27.366 ± 2.22 & 26.229 ± 4.15 & 18.354 ± 2.06 & \textbf{88.465} ± 0.539 & 87.973 ± 1.26 & \textbf{89.056} ± 1.17 & *\textbf{18.333} ± 1.172 & *\textbf{25.418} ± 1.90 & *\textbf{22.707} ± 1.42 \\
        \bottomrule

        \toprule
        \textbf{Personalized} \\
        FedAvg [2017] & 11.133 ± 0.49 & 9.671 ± 0.368 & 13.454 ± 0.87 & 77.039 ± 2.69 & 76.080 ± 1.47 & 75.786 ± 1.76 & 34.625 ± 2.39 & 17.625 ± 0.76 & 17.907 ± 0.24 \\
        
        FedOpt [2021] & 17.417 ± 1.24 & 18.825 ± 0.16 & 15.205 ± 1.07 & 76.073 ± 2.05 &  76.492 ± 1.40 & 76.304 ± 2.76 & 40.275 ± 2.22 & 33.569 ± 1.89 & 43.152 ± 1.43 \\
        
        FedSoft [2022] & 10.814 ± 0.384 & 8.012 ± 0.57 & 8.180 ± 0.01 & 75.547 ± 0.94 & 75.108 ± 1.21 & 73.918 ± 1.41 & 7.812 ± 1.32 & 6.103 ± 0.36  & 3.695 ± 2.52 \\
        
        PACFL [2023] & 13.837 ± 1.72 & 10.454 ± 0.56 & 12.985 ± 1.97 & \textbf{83.805} ± 1.76 & 82.915 ± 1.75 & \textbf{83.722} ± 1.96 & 18.348 ± 0.62 & 20.171 ± 3.49 & 16.266 ± 1.08 \\
        
        CreamFL [2023] & 11.169 ± 2.37 & 11.853 ± 0.815 & 11.338 ± 0.647 & 78.610 ± 0.411 & 78.484 ± 4.85  & 77.008 ± 2.07 & 76.900 ± 3.88 & \textbf{79.461} ± 1.92 & 75.122 ± 3.01 \\
        
        PmcmFL [2024] & 17.010 ± 0.32 & 14.662 ± 1.36 & 14.048 ± 0.06 & 83.031 ± 1.20 & 81.673 ± 1.59 & 79.987 ± 0.82 & - & - & - \\
        \midrule
        \textbf{MMiC} &*\textbf{{19.936}} ± 0.91 & \textbf{18.911} ± 1.13 & \textbf{18.259} ± 1.62 & \textbf{83.576} ± 0.75 & *\textbf{{84.022}} ± 1.19 & \textbf{84.057} ± 1.73 & \textbf{90.726} ± 3.28 & \textbf{75.814} ± 3.35 & \textbf{78.979} ± 2.38 \\
        \textbf{MMiC (MR)} & \textbf{19.492} ± 0.27 & \textbf{19.174} ± 1.32 & \textbf{18.101} ± 1.08 & \textbf{83.519} ± 2.31 & \textbf{83.055} ± 3.74  & 82.948 ± 3.59 & 41.824 ± 6.70 & 33.388 ± 4.52 & 31.489 ± 4.17 \\
        \textbf{MMiC (GF)} & \textbf{19.625} ± 4.797 & *\textbf{{22.796}} ± 6.04 & *\textbf{{18.450}} ± 2.73 & *\textbf{{84.715}} ± 3.13 & \textbf{83.639} ± 2.10 & *\textbf{{84.601}} ± 4.69 & *\textbf{{112.506}} ± 5.10 & *\textbf{{160.99}} ± 6.082 & *\textbf{{124.414}} ± 2.86 \\
        \bottomrule
    \end{tabular}
    }
    \label{tab:overall-performance}
\end{table*}

\subsubsection{\textbf{Datasets, Models and Metrics}}
Our evaluation of MMiC is based on two multimodal classification datasets and one multimodal retrieval dataset.
Because each dataset has distinct characteristics, we followed~\cite{feng2023fedmultimodal,imbalance_reco} in employing different base models, learning configurations and evaluation metrics for each dataset.

\vspace{1mm}
\noindent\textbf{CrisisMMD~\cite{multimodalbaseline2020}.} 
It is a multi-modal dataset for classification consisting of 18,126 samples annotated tweets and images.
We brought in a \textit{\textbf{lightweight multi-modal}} model that used pre-trained embedding layers for input and concatenates shared representations~\cite{gupta2024crisiskan,feng2023fedmultimodal}. Following~\cite{feng2023fedmultimodal}, we employed the \textbf{F1}~\cite{goutte2005probabilistic} score to evaluate the performance of CrisisMMD. 

\vspace{1mm}
\noindent \textbf{UPMC-Food101~\cite{gallo2020image}.} UPMC-Food101 sourced from Food101, containing 101,000 image-text pairs across 101 food categories.
We used \textbf{\textit{CLIP}} with the official ViT version of the backbone pre-trained model~\cite{radford2021learning}. 
We froze the first 11 layers of attention while retaining the last layer of attention and the final output linear layer for parameter updates. 
We employed accuracy to evaluate its performance.

\vspace{1mm}
\noindent \textbf{Flickr30k~\cite{young-etal-2014-image}.} It is an image captioning dataset for retrieval, which is comprised of 31,783 images and the corresponding captions.
We used \textbf{\textit{SCAN}} (Stacked Cross Attention Network)~\cite{lee2018stacked} as the multimodal retrieval model. 
SCAN is a model designed for image-text matching.
We relied on the RSum as the evaluation metric to measure the performance of the retrieval task~\cite{faghri2017vse++}.

\noindent \textbf{Learning settings.} In all of our experiments, we set a global learning rate (0.1, 0.01, 0.05) in the three datasets (CrisisMMD, Food101, Flickr30k) respectively to global optimizers (both $FedAdagrad$ and $FedAdagrad_{MPO}$). Our Parameter difference calculation process is derived from Equation~\ref{eq:adam_efficiency} and we used the Adam optimizer with a learning rate of 0.005 for the clients. These parameters were obtained through our simple tuning in the early stage of our experiment. 
The batch size for the CrisisMMD dataset was set to 15, and 7 for the UPMC-Food101 dataset, 20 for the Flickr30k dataset. All the local epoch was set to 3. 
The loss function on the client was cross-entropy, except for MMiC (GMC-Fusion), where the loss consists of two parts: cross-entropy and contrastive loss between existing modalities~\cite{poklukar2022geometric}.

\subsubsection{\textbf{Benchmark Algorithms}}
We compared MMiC with selected representative works and state-of-the-art works. \textbf{FedAvg~\cite{mcmahan2017communication}} (first work for FL), \textbf{FedOpt~\cite{adaptivefedopt}} (outstanding FL work),  \textbf{FedSoft~\cite{fedsoft}} (a soft SOTA clustered FL), \textbf{PACFL~\cite{vahidian2023efficient}} (an advanced CFL methodology), 
\textbf{CreamFL \cite{yu2023multimodal}} (MFL for addressing missing modality), and \textbf{PmcmFL \cite{baomissing2024}} (a SOTA work for addressing missing modality).

\subsubsection{\textbf{Cluster and Clients Settings}}
\label{sec:settings}
In this experiment, we employed two distance metrics for clustering: Locality Sensitive Hashing (LSH) sketch \cite{yanglsh,liu2024one} and SVD decomposition \cite{vahidian2023efficient}.  All experiments used the LSH
sketch for clustering unless otherwise noted. Each cluster had a test set visible to all clients within the cluster. Additionally, the global server provided accessibility to the test set for all clients.
We simulated the Non-IID scenario by utilizing the \textit{Dirichlet} distribution \cite{teh2010dirichlet} with $\beta_{diri}=0.5$. For all the experiments, we set the number of clients to 50 so that ensured each client had at least 200 training samples.
Our client selection strategy comprised an initial phase of 100 rounds with random selection, followed by a subsequent phase of 200 rounds utilizing Banzhaf client selection (300 global rounds).

\vspace{1mm}
\subsubsection{\textbf{Missing Modality Settings.}}
\label{sec:missing_mdality}
Our setting with missing modalities was governed by two variables: 
\textit{clients with missing modality} (\textbf{mc}) and 
\textit{missing modality rate} (\textbf{mm}), denoting the proportion of all clients selected to have missing modalities and the proportion of missing samples
on these clients, respectively. Our main experiment (Section~\ref{sec:overall-performance}) offered three levels of modality incompleteness settings: (mm@0.2, mc@0.3), (mm@0.2, mc@0.5) and (mm@0.5, mc@0.5). (mm@0.2, mc@0.3) means ratio of mm = 20\% and ratio of mc = 30\%. 
Each client has a fixed number of missing samples, determined by the \textbf{\textit{mm}} rate. In each round, among the clients selected for aggregation, we use \textbf{\textit{mc}} to decide whether the selected clients will activate missing data. Since the selected clients use a random training subset for training in each round, the maximum missing modality rate for a client in that round is $min(\frac{missing\ samples}{subset\ samples}, 1)$.
%

\subsection{Overall Performance}
\label{sec:overall-performance}
We evaluated MMiC based on \textit{global performance} (\textbf{\textit{gbp}}) and \textit{personalized performance} (\textbf{\textit{plp}})
and presented the results in Table~\ref{tab:overall-performance}.
To do the evaluation, we made use of the global test dataset to evaluate the \textit{global} model performance following ~\cite{mcmahan2017communication}, and 
adopted the weighted average performance from client models to evaluate the \textbf{\textit{plp}} following ~\cite{nips_perfedavg_2020}. 
MMiC 
and its two variants secure the top positions across all tasks and all missing modality scenarios.
While MMiC (MR) excelled in classification tasks, being the only algorithm to achieve an F1 score exceeding 30 on CrisisMMD (mm@0.2, mc@0.5) at \textbf{\textit{gbp}}. MMiC (GF) significantly outperformed all other algorithms in modality retrieval tasks across all three metrics. These results highlighted the superiority of MMiC in handling diverse missing modality scenarios.

FedOpt demonstrated strong \textbf{\textit{gbp}} on the CrisisMMD dataset (mm@0.2, mc@0.3) and (mm@0.2, mc@0.5). 
FedSoft's performance is medium across experiments with multiple missing modalities settings and datasets both in global and personalized evaluations. As a CFL algorithm, PACFL do not exhibit outstanding performance in \textbf{\textit{gbp}}. In contrast, it demonstrated excellent performance in personalized evaluation. 
CreamFL is primarily an algorithm designed for retrieval tasks. 
The overall performance of CreamFL is relatively mediocre except for Flickr30k (mm@0.2, mc@0.5), their method ranked second in this setting both in global and personalized evaluations.
PmcmFL currently supports only classification tasks; however, its performance in classification tasks is somewhat limited. Overall, MMiC consistently outperforms traditional methods for addressing missing modality challenges in FL.

\subsection{Ablation Study}
\label{sec:ablation_study}
We conducted an ablation study on the CrisisMMD, Food101, and Flickr30k datasets under the settings of mm@0.2 and mc@0.5 (Table~\ref{tab:ablation_combined}). The ablation modules are denoted as `b', `p', and `s' (BPI, MPO and PPS respectively). It shows that MMiC consistently achieves higher accuracy in the final rounds across multiple ablation tests. Specifically, `w/o b' and `w/o s' saw a decrease of more than 5\% compared to MMiC in CrisisMMD while others still performed poorly. The final performance of `w/o p' is close to MMiC, yet it remained approximately 2\%, 0.2\% and 3\% lower than MMiC in three datasets respectively. 
In addition, removing any of the two contributions leads to varying degrees of performance decline, with the resulting performance consistently falling between that of MMiC and the configuration with a single module removed in Food101 and Flickr30k. 
We conclude that removing any module leads to performance degradation. We further proved the effectiveness of BPI in Section~\ref{apx:exp-banzhaf} and MPO in Section~\ref{apx:exp_mpo}. We believe that BPI helps MMiC by reducing the weight of clients heavily affected by missing modalities, ensuring well-performing clients are prioritized. MPO further improves aggregation by minimizing the influence of high-risk clusters with missing modalities.
\vspace{-2mm}
\begin{table}[htbp]
\centering
\caption{Ablation study on CrisisMMD, Food101 and Flickr30k under (mm@0.2, mc@0.5). 
}
\label{tab:ablation_combined}
\resizebox{\linewidth}{!}{%
\begin{tabular}{lc|c|c}
\toprule
\textbf{Configuration}& \textbf{CrisisMMD (Acc)}  & \textbf{Food101 (Acc)} & \textbf{Flickr30k (RSum)} \\
\midrule
w/o (b, p, s)& 57.22 & 86.97 & 4.28 \\
w/o s        & 58.91 & 88.54 & 6.99 \\
w/o p        & 61.07 & 88.91 & 8.18 \\
w/o b        & 57.03 & 87.26 & 8.10 \\
only s       & 60.36 & 88.03 & 7.08 \\
only p       & 59.87 & 87.49 & 4.84 \\
only b       & 61.24 & 87.41 & 5.36 \\
MMiC         & \textbf{63.55} & \textbf{90.12} & \textbf{11.02} \\
\bottomrule
\end{tabular}
}
\end{table}

\subsection{Experiments on other modalities}
\label{apx:othermodality}
In this section, we further broaden our experimental evaluation by incorporating more modalities, aiming to investigate the feasibility of our algorithm on non-dominant modalities, aiming to show that MMiC is modality-agnostic.
UCI-HAR~\cite{uci-har} and KU-HAR~\cite{ku-har} both encompass smartphone sensor data utilized for the classification of diverse human activities. They include accelerometer and gyroscope readings gathered from a multitude of participants. 
We compared MMiC with FedOpt, which exhibited the good performance 
in Section \ref{sec:overall-performance}
, for this analysis. 
The experimental setup involved 30 clients running 150 global rounds, with all other parameters remaining consistent with those outlined in Section~\ref{sec:settings}. 
%
The findings of this experiment are presented in Table~\ref{tab:extend_modalities}.
MMiC demonstrated a pronounced superiority over FedOpt, with enhancements of 2\% and 6\% in the KU-HAR and UCI-HAR respectively. 
\begin{table}[h!] \small
\centering
\caption{Performance comparison of MMiC and FedOpt on UCI-HAR and KU-HAR datasets.}
\begin{tabular}{lcc}
\toprule
\textbf{Dataset/Algorithm} & \textbf{MMiC (\%)} & \textbf{FedOpt (\%)} \\
\midrule
UCI-HAR (Acc) & \textbf{86.99} & 75.83 \\
KU-HAR (Acc) & \textbf{80.89} & 73.24 \\
\bottomrule
\end{tabular}
\label{tab:extend_modalities}
\end{table}


\begin{figure}
	\centering
        \includegraphics[width=\linewidth]{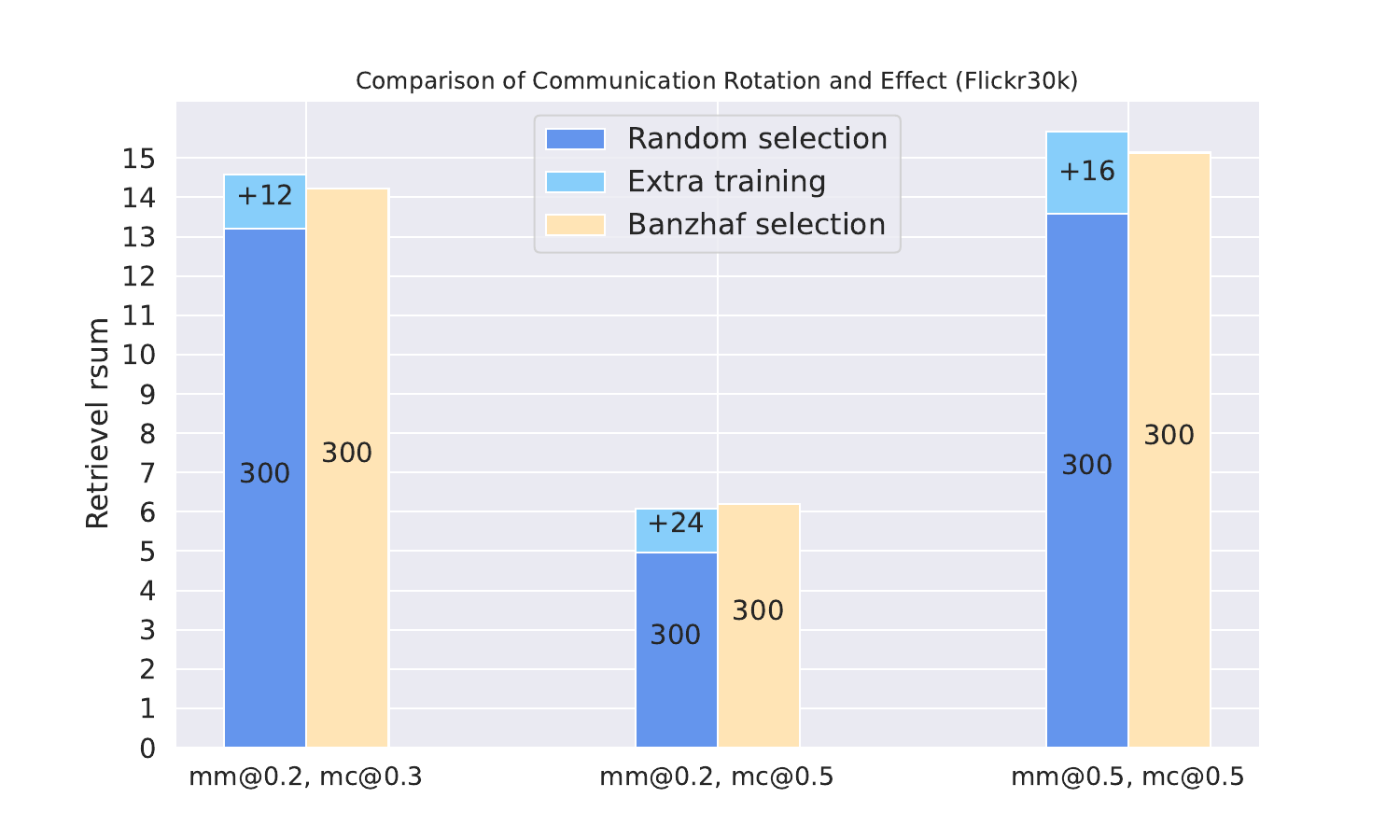}
        \caption{The number of additional training rounds needed for the Random Client Selection to achieve the same performance as the Banzhaf Client Selection (Flickr30k). The blue bar means random selection, and the yellow represents the BPI selection.}
        \label{fig:banzhaf-selection}
        \vspace{-5mm}
\end{figure}

\subsection{Effectiveness of Banzhaf Client Selection}
\label{apx:exp-banzhaf}
The federated architecture had shown the feasibility of client selection, as demonstrated in~\cite{nishio2019client, cho2020client}. In this section, we compared training efficiency under missing modalities and show that BPI achieves higher efficiency than random selection. The Banzhaf client selection strategy aimed to optimize global aggregation efficiency by minimizing communication rounds.
Figure~\ref{fig:banzhaf-selection} illustrates 
the comparison of communication rounds and performance between the setting with Banzhaf client selection and random client selection across three levels of missing modality. 
We used Flickr30k dataset in this evaluation. It should be noted that randomly selected patterns required additional (12, 24, or 16) training rounds to surpass or approach the performance of the Banzhaf client selection. To some extent, MMiC facilitates faster convergence of the global model under missing modality settings by prioritizing the selection of core clients.


\subsection{MPO: A Potent Solution against Over-fitting}
\label{apx:exp_mpo}
\begin{figure}[htbp]
	\centering
        \includegraphics[width=\linewidth]{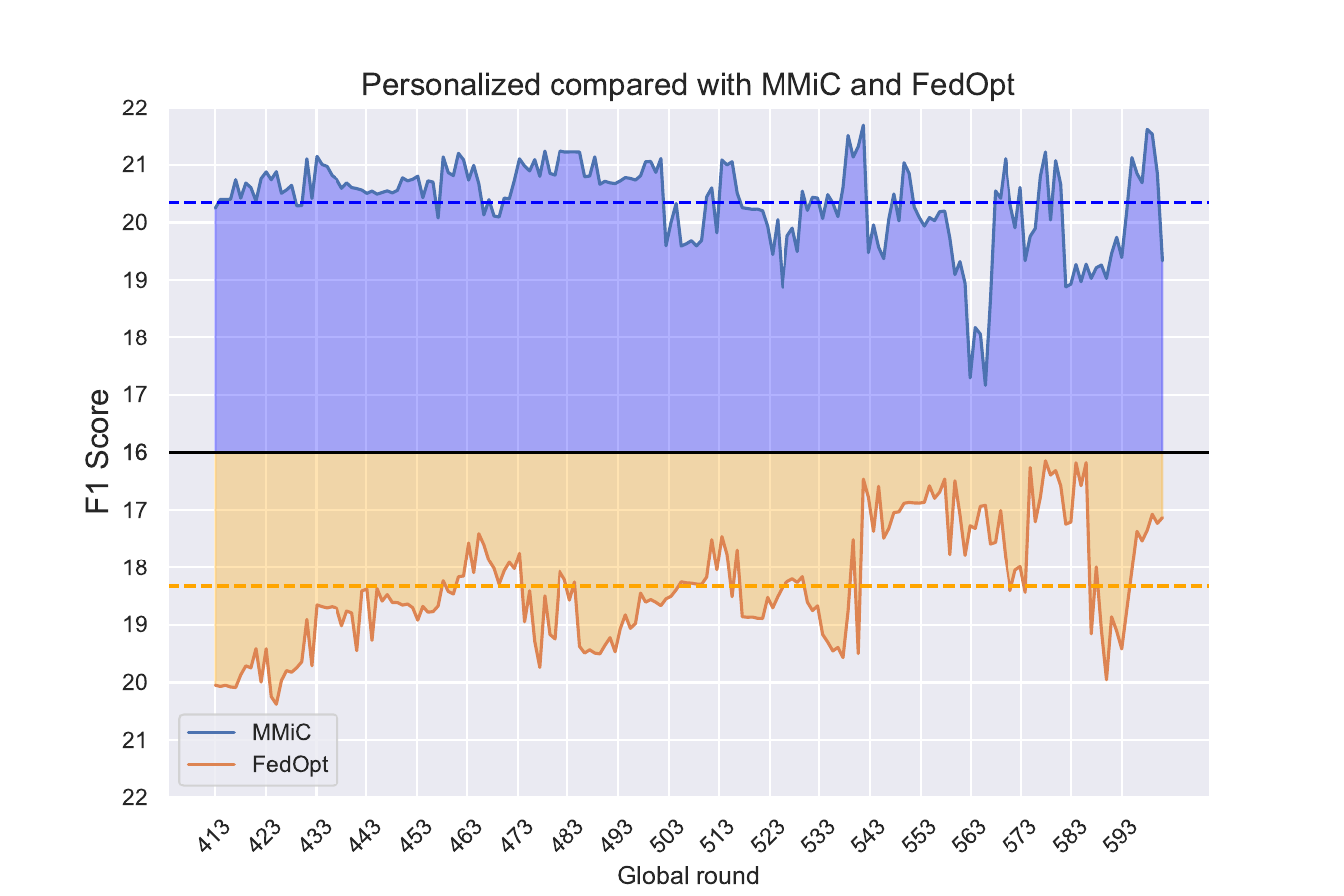}
        \caption{Personalized performance of comparison after continuous training since the model convergence (400 rounds). The two dotted lines are the average scores. The upper part (blue) represents $FedAdagrad_{MPO}$ (MMiC) and the lower part (yellow) represents $FedAdagrad$ (FedOpt). 
        }
        \label{fig:optimal_overfit}
        \vspace{-4mm}
\end{figure}

In this section, we show additional features of MPO against overfitting. We only implemented MMiC with MPO and deactivated PPS and BPI. 
Therefore, the comparison is between $FedAdagrad_{MPO}$ (MMiC) and $FedAdagrad$ (FedOpt) on dataset CrisisMMD (mm@0.2, mc@0.3) 
. Our purpose is to evaluate how the dynamic updating of parameters in $FedAdagrad_{MPO}$ contributes to enhanced stability. During the training, it had been converged after 400 rounds. We continued the training after 400 rounds while observing the personalized performance of $FedAdagrad$ and 
$FedAdagrad_{MPO}$, which showed the stability of our algorithm. 
The personalized evaluation is based on the average weighted performance of each client. It can be observed that the personalized performance of FedOpt begins to decline after round 413 and never recovers to its peak level thereafter. This indicated that some clients have already experienced overfitting, leading to a degradation in performance. In contrast, although the performance curve of MMiC experiences a short decline after round 543 as shown in Figure~\ref{fig:optimal_overfit}. 
$FedAdagrad_{MPO}$ forms risk by calculating the covariance of historical multi-round returns between pairs of clients (Equation~\ref{eq:cov_risk}). When partial clients overfit, the covariance increases, leading to a higher risk and an overall increase in RARC. When the majority of clients overfit, the total return decreases, yet the overall RARC still increases. An increase in RARC indicates that fewer parameters will be aggregated to the global model.
The experimental results demonstrated that the stability of our algorithm is significantly enhanced by incorporating the MPO, particularly in mitigating performance degradation caused by overfitting during the training of clients.
\vspace{-0.5mm}

\begin{figure*} [!htbp]
    \centering
    \includegraphics[width=0.24\linewidth]{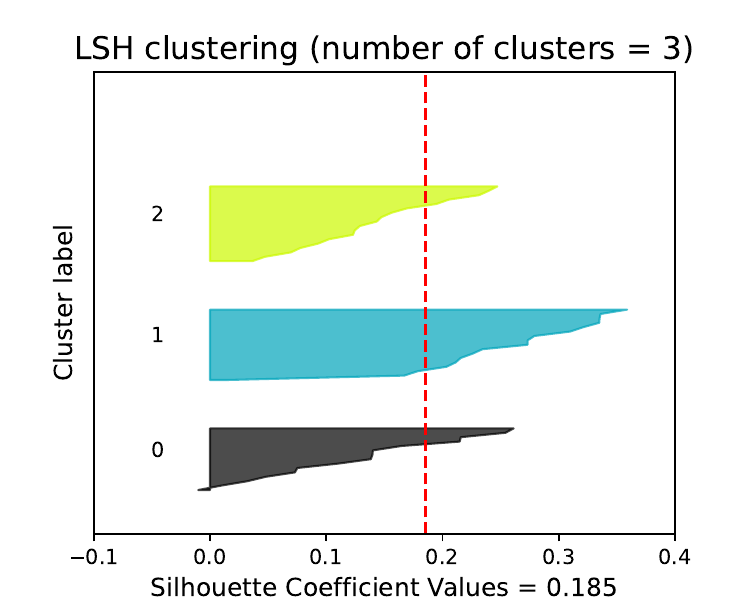}
    \hspace{0.02\linewidth}
    \includegraphics[width=0.24\linewidth]{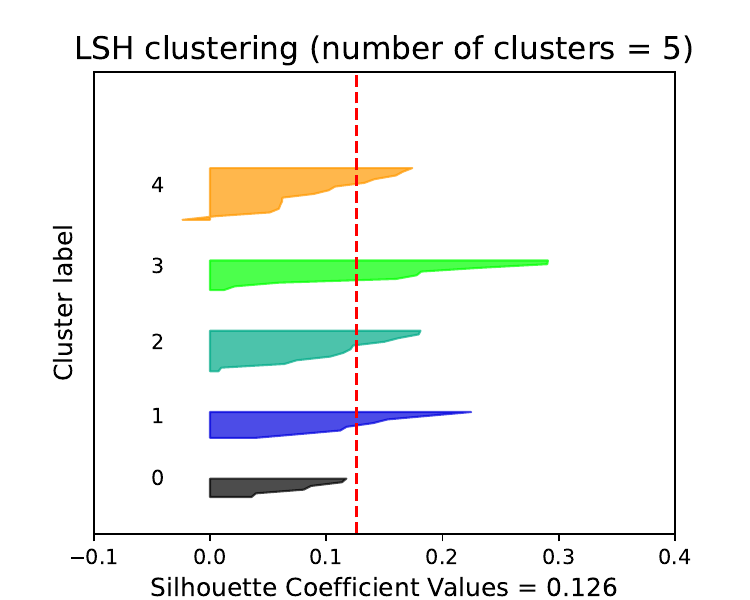}
%
    \includegraphics[width=0.24\linewidth]{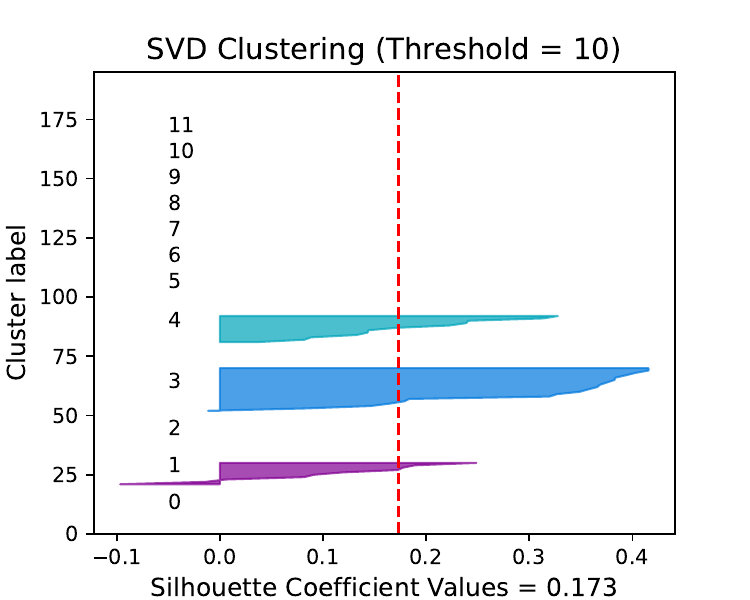}
    \includegraphics[width=0.24\linewidth]{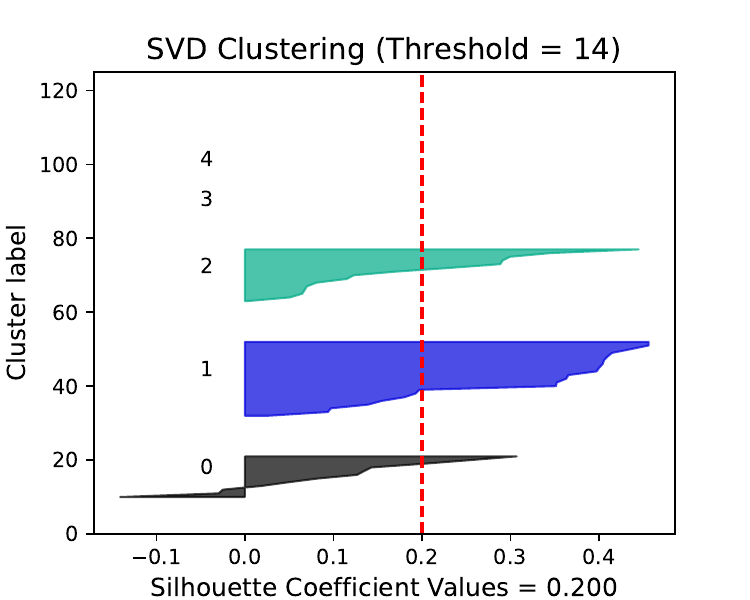}
    \caption{
    The two plots on the left depict the SCV of LSH clustering under two different hyperparameter (number of clusters), while the two plots on the right show the SCV of SVD-based clustering with the threshold hyperparameter. The numbers within the images represent the indices of the cluster labels.}
    \label{fig:lsh-silhouette-coefficient}
\end{figure*}

\subsection{Global Performance Influenced by Clustered Federated Algorithms}
\label{apx:performance_by_cfl}

We explored the impact of clustering quality on MMiC performance. Figure~\ref{fig:lsh-silhouette-coefficient} plots \textit{Silhouette Coefficient Value (SCV)} of two clustering algoritms with two different hyperparameters respectively. This experiment was conducted on dataset CrisisMMD (mm@0.2, mc@0.5)
and we investigated how to change clustering hyperparameters 
to get a higher SCV. The SCV quantifies the ratio of the intra-cluster distance, defined as the distance from the sample point to the nearest cluster to which it doesn't belong~\cite{shahapure2020cluster}. \textbf{A larger SCV} signifies a greater separation between the cluster structure containing the sample point and its nearest neighbouring cluster structure, thereby indicating \textbf{superior clustering quality} under these conditions.  For Locality Sensitive Hashing (LSH) 
clustering, their clustering algorithm is based on KMeans~\cite{kmeans2002}, which requires specifying the number of clusters. We adjusted the number of clusters from 1 to 20 to identify the two highest SCV. When the number of clusters is set to 3, the SCV attained its peak at 0.185. With 7 clusters and its second highest value of 0.125.
For Singular Value Decomposition (SVD) 
clustering, which employs hierarchical clustering~\cite{johnson1967hierarchical}, we varied the threshold from 1 to 20 to determine the optimal thresholds, identifying two values: 10 (SCV = 0.18) and 14 (SCV = 0.2). This preliminary selection of clustering parameters assisted us in identifying better and more suitable clustering algorithms and parameters for MMiC.

\begin{figure}[!htbp]
    \centering
    \includegraphics[width=1.0\linewidth]{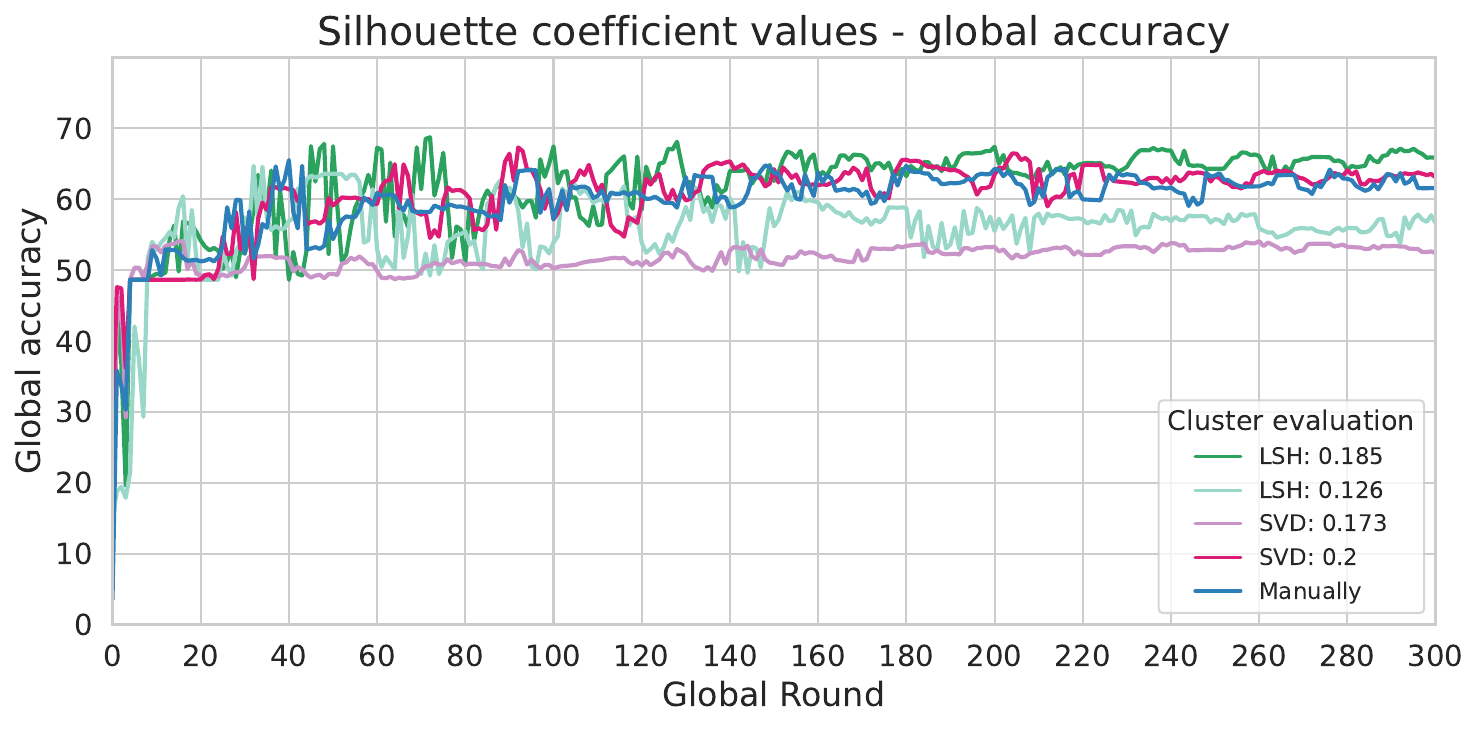}
    \caption{We compared two clustering algorithms, LSH and SVD, by selecting the top-2 SCV hyperparameters for each and evaluating their global performance on MMiC. The blue curve is manual clustering.}
    \label{fig:comparision-clusters}
    \vspace{-4mm}
\end{figure}

Figure~\ref{fig:comparision-clusters} investigated the selection process of our clustering strategies. What we found are: (1) It's shown that Poverty Parameter Substitution within clusters heavily relies on the quality of client clustering. For instance, in LSH clustering, a higher clustering quality (a higher SCV) often correlates with better global performance (LSH: 0.185 compared with LSH: 0.126). (2) By comparing different clustering methods, we can infer the contribution of each clustering algorithm to MMiC based on global performance. It is evident that LSH clustering outperformed SVD 0.2 in terms of global model performance when the SCV reached at 0.185. (3) We included manual clustering as a gold standard 
(clustering based on the overlap of client labels). Overall, both LSH (0.185) and SVD (0.2) clustering achieved global model performance comparable to that of manual clustering, making them reliable references for the clustering algorithms employed in this study. When compared to SVD, LSH demonstrated slightly superior performance.

\subsection{Time Complexity of MPO}
\label{apx:time_analysis}
Since the Banzhaf Client Selection do not incur an additional cost, the time complexity of MMiC compared to other works depended on the complexity of Markowitz Portfolio Optimization. The aggregation process of $FedAdagrad_{MPO}$ involves the computation of RARC in each cluster, which is then uploaded to the server. The server utilizes the RARC of each cluster to dynamically adjust the beta parameter for aggregation. Since the computational complexity of beta is linear and negligible, the overall complexity of MPO is primarily dominated by the RARC computation. The time complexity of MPO is $O(\frac{K^2}{M} \cdot T + K \cdot T)$. Here we deduce:

\begin{itemize}
    \item We got average \( \frac{K}{M} \) clients for each cluster. 
    Time complexity of \(\sigma_{ij}\) was \(O(T)\).
    
    Time complexity of risk $\bar{\sigma}$: \[
    \bar{\sigma} = \sum_{i=1}^{\frac{K}{M}} \sum_{j=1}^{\frac{K}{M}} w_i w_j \sigma_{ij}.
    \] could be obtained from:
    \[
    O\left( (\frac{K}{M})^2 \right) \times O(T) = O\left((\frac{K}{M})^2 \cdot T\right)
    \]
    In the covariance $\sigma_{ij}$ section, we can employ the concept of \textit{prefix sums}~\cite{prefix_sums} to efficiently store historical data. The time complexity of constructing prefix sums was extra $O(\frac{K^2}{M^2})$. Hence this part could be improved to $O(\frac{K^2}{M^2}) + O(\frac{K^2}{M^2})$.
    \item Return $A$ computation is to do the weighted sum for each cluster, and the time complexity of $A$ is $O(\frac{K}{M})$ to each cluster.
    \item $M$ clusters will upload their RARC to the server and calculate the weighted sum:
    \[
    \left( O(\frac{K^2}{M^2} + \frac{K}{M})+ O(\frac{K^2}{M^2}) \right) \times M  = O(\frac{K^2}{M} + K) + O(\frac{K^2}{M})
    \]
    Meanwhile, MMiC has a total $T$ global rounds. As \textit{prefix sums} only build at once, we got:
    \[
     O(\frac{K^2}{M} + K)  \times T+ O(\frac{K^2}{M}) = O\left(\frac{K^2 \cdot T}{M} + K \cdot T + \frac{K^2}{M}\right) 
    \]
    $O(\frac{K^2}{M})$ was less than $O(\frac{K^2 \cdot T}{M})$ apparently. Therefore, the time complexity of MPO is $O(\frac{K^2}{M} \cdot T + K \cdot T)$.
\end{itemize}

\section{Conclusions}
In this work, we introduce MMiC, a model-independent and modality-agnostic framework designed to mitigate the impact of missing modalities in multimodal federated learning. By substituting parameters within clustered federated learning, adapting the Banzhaf Power Index for client selection, and extending Markowitz Portfolio Optimization for global aggregation, MMiC demonstrates improved performance over existing federated architectures in addressing missing modalities. Moving forward, we plan to further enhance the framework’s robustness by substituting additional poverty parameters and exploring more efficient solutions.

\section{GenAI Usage Disclosure}
The authors acknowledge the use of an AI language model to assist in the linguistic refinement of selected sections of this paper, specifically the Introduction and Methodology. All the wordings, equations, and conclusions were revised again after being reviewed by the authors.
\bibliographystyle{ACM-Reference-Format}
\bibliography{references}

\end{document}